# Support and Plausibility Degrees in Generalized Functional Models

Paul-André Monney
University of Fribourg, Seminar of Statistics
Beauregard 11-13
CH-1700 Fribourg, Switzerland
paul-andre.monney@unifr.ch

## Abstract

By discussing several examples, the theory of generalized functional models is shown to be very natural for modeling some situations of reasoning under uncertainty. A generalized functional model is a pair $(f, P)$ where $f$ is a function describing the interactions between a parameter variable, an observation variable and a random source, and $P$ is a probability distribution for the random source. Unlike traditional functional models, generalized functional models do not require that there is only one value of the parameter variable that is compatible with an observation and a realization of the random source. As a consequence, the results of the analysis of a generalized functional model are not expressed in terms of probability distributions but rather by support and plausibility functions. The analysis of a generalized functional model is very logical and is inspired from ideas already put forward by R.A. Fisher in his theory of fiducial probability.

## 1 The Theories of Bayes and Fisher

Jessica is a young woman suspecting that she might be pregnant. To find out about her status she decides to go to the local pharmacy and buys a pregnancy test. The test result indicates that she is not pregnant, but she knows that such tests are not fully trustworthy. The question is then how to evaluate the chance that she is pregnant in spite of the negative test result.

Of course, one possibility is to use the Bayesian theory to analyze the situation. The Bayesian model is as follows. Let $\theta$ denote the variable indicating her true pregnancy status. The set of possible values of $\theta$ is $\Theta = \{-1, +1\}$ where $-1$ means that she is not pregnant and $+1$ means that she is pregnant. Similarly, let $\xi$ denote the variable indicating a test result. The set of possible values of $\xi$ is $X = \{-1, +1\}$ where $-1$ represents a negative test result and $+1$ a positive test result. The reliability of the test can be expressed by two numbers, namely the chance $p$ that the test will indicate a negative result when a woman in not pregnant, and the chance $p'$ that the test will indicate a positive result when a woman is pregnant. It is reasonable to assume that both $p$ and $p'$ are rather high. This information is represented by the conditional probabilities

$$P(\xi = -1|\theta = -1) = p, \; P(\xi = 1|\theta = 1) = p'. \quad (1)$$

Suppose that Jessica's prior about her status is given by the probabilities

$$P(\theta = -1) = y, \; P(\theta = +1) = 1 - y. \quad (2)$$

Then by Bayes theorem the posterior probability that Jessica is not pregnant is

$$P'(\theta = -1) = \frac{yp}{yp + (1-y)(1-p')}. \quad (3)$$

According to the Bayesian theory, this represents Jessica's degree of confidence in the fact that she is not pregnant.

Now assume that the probability of getting a correct negative result is the same as getting a correct positive result, i.e. $p = p'$. This means that the test reveals the true pregnancy status with probability $p$. This probability is an indicator of the confidence in the test result. Then by formula (3) the posterior probability that Jessica is not pregnant is

$$P'(\theta = -1) = \frac{yp}{yp + (1-y)(1-p)}. \quad (4)$$

This is the Bayesian solution to Jessica's pregnancy test problem. However, Jessica is unable to give the prior probability of her being pregnant because she is not comfortable with the idea of giving a precise number to estimate this chance. As can be seen in formula (4), without a prior it is impossible to find the posterior probability of her not being pregnant. The Bayesian theory requires prior probabilities to compute posterior probabilities. But Jessica is still interested in finding a numerical value expressing the chance of her not being pregnant considering the negative test result. How can we find such a numerical value ? In 1930, R.A. Fisher identified a class of problems in which it appeared to him that inductive probability statements could legitimately be made without prior probabilities being used. It turns out that Jessica's problem when $p = p'$ is one of them. In this example, Fisher's reasoning would go as follows. First define the variable

$$\omega = \theta \cdot \xi. \qquad (5)$$

The distribution of $\omega$ is independent of $\theta$ as the following development shows. First we have

$$\begin{aligned} P(\omega = 1) &= P(\omega = 1|\theta = -1)P(\theta = -1) \\ &\quad + P(\omega = 1|\theta = 1)P(\theta = 1). \end{aligned} \qquad (6)$$

But by equation (5) if $\theta = -1$ then $\omega = -\xi$ and if $\theta = 1$ then $\omega = \xi$. Therefore

$$\begin{aligned} P(\omega = 1) &= P(-\xi = 1|\theta = -1)P(\theta = -1) \\ &\quad + P(\xi = 1|\theta = 1)P(\theta = 1) \\ &= p \cdot P(\theta = -1) + p \cdot P(\theta = 1) \\ &= p \end{aligned} \qquad (7)$$

and similarly $P(\omega = -1) = 1 - p$. Fisher called such a variable $\omega$ a pivotal quantity or a pivot. Generally speaking, a pivotal quantity is a function of $\theta$ and $\xi$ whose distribution is independent of $\theta$. The observation $\xi = -1$ Jessica makes doesn't permit her to infer some information on $\omega$ because both $\omega = +1$ and $\omega = -1$ are compatible with the observation (see equation (5)). So after observing $\xi = -1$ we still have $P(\omega = 1) = p$ and $P(\omega = -1) = 1 - p$. These probability statements are as valid as before observing $\xi = -1$. The distribution of $\omega$ can be seen as a feature of the test device: $P(\omega = 1) = p$ means that the probability of getting a correct test result is $p$ because $\omega = 1$ when $\theta$ and $\xi$ indicate the same pregnancy status. Under the assumption that $\omega = 1$, the observation $\xi = -1$ logically implies that $\theta = -1$ since $\omega = \theta \cdot \xi$. Since the assumption that $\omega = 1$ is true with probability $p$, it follows that this probability can be transfered to its logical consequence $\theta = -1$, i.e. $P(\theta = -1) = p$. In a similar way, it can be derived that $P(\theta = 1) = 1 - p$. This argumentation is an instance of Fisher's fiducial theory. The probabilities $P(\theta = 1) = 1 - p$ and $P(\theta = -1) = p$ represent the so-called fiducial distribution of $\theta$. Fisher noticed that the fiducial distribution coincides with the Bayesian posterior if a uniform prior distribution is assumed for the variable $\theta$, but Jessica doesn't have any prior (not even the uniform one) and so the statement $P(\theta = -1) = p$ cannot be justified by Bayes' theorem.

There are situations where an observation does provide information about a pivotal quantity $\omega = g(\theta, \xi)$. This information is represented by the set $\Omega'$ of all $\omega \in \Omega$ that are compatible with the observation. The elements in $\Omega - \Omega'$ are impossible in view of the observation. This information is taken into account by conditioning the initial distribution of $\omega$ on the set $\Omega'$. This idea of conditioning the distribution of the pivot is also present in Fisher's fiducial theory. In the fiducial theory, using the pivot $\omega = g(\theta, \xi)$ and the observation $\xi = x$, every element $o \in \Omega'$ logically implies that $\theta = t$ for some $t \in \Theta$. If $P'$ denotes the conditional distribution of $\omega$ given $\Omega'$, then the fiducial probability of $t$ is given by the probability of all $o \in \Omega'$ that logically imply $\theta = t$, i.e.

$$P_{fid}(t) = \sum \{P'(o) : o \in \Omega',\ o \text{ implies } t\}. \qquad (8)$$

## 2  Generalized Functional Models

In this section we consider models where the spirit and the logic of Fisher's fiducial argument can be applied. However, the theory presented here is not the fiducial theory itself. Within the fiducial theory, the reasoning starts with two variables: the parameter variable $\theta$ and the observation variable $\xi$. Then one tries to find a pivotal quantity $\omega = g(\theta, \xi)$ from which the fiducial distribution of $\theta$ is determined using the observation.

In generalized functional models we start from three variables: the parameter and observation variables $\theta$ and $\omega$ plus a third variable $\omega$. The variable $\omega$ is a random variable whose distribution $P$ is known and assumed to be independent of $\theta$. This is a major difference with the fiducial theory because the fiducial theory requires finding a function of $\theta$ and $\xi$ whose distribution can be determined and proved to be independent of $\theta$. In the fiducial theory, conditional distributions form the initial knowledge before the observations are made. However, in generalized functional models the initial knowledge is not given by such conditional distributions, but rather by the distribution of $\omega$ assumed to be independent of $\theta$, and by a function of the variables $\theta$ and $\omega$ to be described below.

Now generalized functional models are precisely defined. Let $\Omega$ denote the set of all possible values of the variable $\omega$ and as usual let $\Theta$ and $X$ denote the set of possible values of $\theta$ and $\xi$ respectively. It is assumed that there is exactly one correct but unknown value of $\theta$. Let $t^*$ be this value. The problem is to generate knowledge about $t^*$ in the light of the available information. The interaction between the three variables $\omega, \xi$ and $\theta$ is represented by a function

$$f : \Theta \times \Omega \to X \qquad (9)$$

that is determined from the situation under investigation (see the examples in section 4). The value $f(t, o)$ specifies what would necessarily be observed if $t$ was the correct parameter value and $o$ was the outcome of the random variable $\omega$. It is important to remark that the function $f$ is completely determined from the logic of the situation under investigation. Thus a generalized functional model is a pair $(f, P)$ where $f$ is a function as in (9) and $P$ is a known probability distribution on $\Omega$ that is independent of $\theta$.

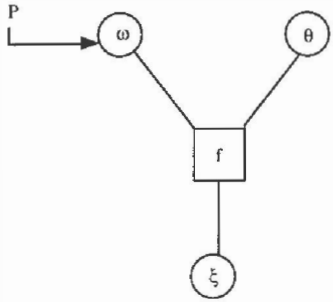

This figure shows that $f$ describes a relation between the variables $\omega, \theta, \xi$ and $P$ is a probability distribution on $\Omega$. In a generalized functional model $(f, P)$, suppose that the value of the variable $\xi$ is observed to be $x$, i.e. $\xi = x$. For the time being, assume that this observation is completely compatible with all possible values of $\omega$, i.e.

$$\{o \in \Omega : \exists\, t \in \Theta \text{ such that } f(t, o) = x\} = \Omega. \quad (10)$$

This means that the observation $x$ could have been generated by any value $o \in \Omega$. In such a situation it is natural to consider that the distribution $P$ of $\omega$ remains the same after observing $x$. The same idea is present in the fiducial theory. Under the assumption that the outcome of the random variable $\omega$ was $o$, the observation $x$ logically implies that the correct value of $\theta$ must be in the set

$$\Gamma_x(o) = \{t \in \Theta : f(t, o) = x\}. \quad (11)$$

So the value $o$ is an argument in favor of the hypothesis that $t^*$ is in $\Gamma_x(\omega)$ because if $o$ was the outcome then the hypothesis would necessarily be true. Since $o$ was the outcome of $\omega$ with probability $P(o)$, it follows that $o$ supports the hypothesis that $t^* \in \Gamma_x(o)$ to the degree $P(o)$. This reasoning is essentially the same as the one used to define the fiducial distribution of $\theta$ when an observation is made. However, in the fiducial theory and in the functional models traditionally considered, the sets $\Gamma_x(o)$ are always assumed to be singletons, which is not necessarily the case in this paper (see sections 3 and 4 below). This is why the functional models considered in this paper are called generalized functional models and not simply functional models. Classical, non-generalized, functional models have been studied by Fraser [3], Bunke [1], Plante [5] and Dawid, Stone [2]. See also Shafer [7].

Now consider the case where the observation $x$ does provide some knowledge about $\omega$. This happens when the set

$$v_x(\Omega) = \{o \in \Omega : \exists\, t \in \Theta \text{ such that } f(t, o) = x\} \quad (12)$$

is different from $\Omega$. Only the elements of $v_x(\Omega)$ are possible values for the outcome of the random variable $\omega$ in view of the observation $x$. So the initial probability distribution $P$ of $\omega$ must be conditioned on $v_x(\Omega)$. This defines a new probability measure $P'$ on $v_x(\Omega)$. Under the assumption that $o \in v_x(\Omega)$ was the outcome of $\omega$, the observation $x$ implies that $t^*$ is in the set

$$\Gamma_x(o) = \{t \in \Theta : f(t, o) = x\}. \quad (13)$$

Since $o$ was the outcome of $\omega$ with probability $P'(o)$, it follows that $o$ in $v_x(\Omega)$ supports the hypothesis that $t^*$ is in $\Gamma_x(o)$ to the degree $P'(o)$. Remark that $\Gamma_x(o)$ is not empty for all $o \in v_x(\Omega)$.

## 3    Generalized Functional Models and Hints

The goal of this section is to show how the knowledge about $\theta$ generated by the observations in a generalized functional model can be expressed with the theory of hints, which is a theory strongly related to the Dempster-Shafer theory of evidence [4], [6]. From a generalized functional model $(f, P)$, an observation $x$ generates the set $v_x(\Omega)$ of values of $\omega$ that are compatible with $x$. This knowledge is used to condition the probability measure $P$ on $v_x(\Omega)$ and let $P'$ denote the resulting probability measure. Using the functional relation $f$ and the observation $x$, each possible outcome $o \in v_x(\Omega)$ logically implies that $t^*$ is in the subset $\Gamma_x(o)$ of $\Theta$. This defines the mapping

$$\Gamma_x : v_x(\Omega) \to 2^\Theta. \quad (14)$$

Since $o$ is the outcome of the random variable $\omega$ with probability $P'(o)$, it follows that $o$ supports the hypothesis that $t^* \in \Gamma_x(o)$ to the degree $P'(o)$. Putting things together, the observation $x$ and the generalized functional model $(f, P)$ define the object

$$\mathcal{H}(x) = (v_x(\Omega), P', \Gamma_x, \Theta) \quad (15)$$

which is called a *hint*. It represents the knowledge about $\theta$ generated by the observation $x$. A hint $\mathcal{H}(x)$ generates two functions called the support and plausibility functions. The support function is constructed as follows. For an arbitrary subset $H \subseteq \Theta$, consider the hypothesis that $t^*$ is in $H$. This hypothesis is either true or false, but to what extend is it supported by the hint $\mathcal{H}(x)$? If the outcome of $\omega$ is in the set

$$u_x(H) = \{o \in v_x(\Omega) : \Gamma_x(o) \subseteq H\}, \quad (16)$$

then the hypothesis $t^* \in H$ is definitely true because if the outcome is an element $o$ in $u_x(H)$ then $t^* \in \Gamma_x(o)$ and $\Gamma_x(o) \subseteq H$, which implies that $t^* \in H$. Since the outcome of $\omega$ is in $u_x(\Omega)$ with probability $P'(u_x(H))$, it is natural to call this probability the degree of support of the hypothesis $t^* \in H$ (in the rest of the paper, such a hypothesis will simply be denoted by $H$). The degree of support of the hypothesis $H$ is written $sp(H)$, i.e.

$$sp(H) = P'(u_x(H)). \quad (17)$$

The corresponding function $sp : 2^\Theta \to [0, 1]$ is called a support function.

It is also interesting to look at the degree of compatibility of the hypothesis $H$ with the knowledge about $\theta$ represented by the hint $\mathcal{H}(x)$. An element $o \in v_x(\Omega)$ is compatible with the hypothesis $H$ if it is possible that $o$ is the outcome of $\omega$ and at the same time $H$ is true. But this is the case when $\Gamma_x(o) \cap H \neq \emptyset$. So

$$v_x(H) = \{o \in v_x(\Omega) : \Gamma_x(o) \cap H \neq \emptyset\} \quad (18)$$

is the set of elements that are compatible with the hypothesis $H$. Since the probability that the outcome of $\omega$ is compatible with $H$ is $P'(v_x(H))$, it is natural to define this probability as the degree of compatibility, or the degree of plausibility, of the the hypothesis $H$. It is written $pl(H)$, i.e.

$$pl(H) = P'(v_x(H)). \qquad (19)$$

This can be done for all subsets $H \subseteq \Theta$ and the corresponding function $pl : 2^\Theta \to [0,1]$ is called a plausibility function. Elementary properties of support and plausibility functions are listed below:

1. $sp(\emptyset) = 0$, $sp(\Theta) = 1$
2. $pl(\emptyset) = 0$, $pl(\Theta) = 1$
3. For all $H \subseteq \Theta$, $pl(H) = 1 - sp(H^c)$, $sp(H) \leq pl(H)$
4. If $H \subseteq H'$, then $sp(H) \leq sp(H')$ and $pl(H) \leq pl(H')$
5. Let $H_i \subseteq \Theta$, $i = 1, \ldots, n$ be a collection of subsets of $\Theta$. If we define $S = \{I \subseteq \{1, \ldots, n\} : I \neq \emptyset\}$ then

$$sp(\cup_{i=1}^n H_i) \geq \sum \{(-1)^{|I|+1} sp(\cap_{i \in I} H_i) : I \in S\}.$$

Property 5 shows that general support functions are not additive measures, e.g. $sp(H) + sp(H^c) \leq 1$. Remark also that the inequality of property 5 would be transformed into an equality if $sp$ was replaced by a probability measure. For simplicity, the abbreviations $sp(t)$ and $pl(t)$ will be used instead of $sp(\{t\})$ and $pl(\{t\})$.

Support functions are known as belief functions in the Dempster-Shafer theory of evidence [6]. A function

$$Bel : 2^\Theta \to [0,1] \qquad (20)$$

is a belief function if there exists a function $m : 2^\Theta \to [0,1]$ satisfying

$$m(\emptyset) = 0$$
$$\sum \{m(X) : X \subseteq \Theta\} = 1 \qquad (21)$$

and such that

$$Bel(H) = \sum \{m(X) : X \subseteq H\} \qquad (22)$$

for all $H \subseteq \Theta$. The support function $sp$ associated with the hint $\mathcal{H}(x)$ is a belief function whose corresponding $m$ function is given by

$$m(X) = \sum \{P'(o) : \Gamma_x(o) = X\}. \qquad (23)$$

There is a special kind of hints called precise hints. A hint $\mathcal{H}(x)$ is called precise when $\Gamma_x(o)$ is a singleton for all elements $o \in v_x(\Omega)$. For precise hints, the corresponding support and plausibility functions coincide and they are probability measures, i.e. $sp(H) = pl(H)$ for all $H \subseteq \Theta$ and $sp$ is a probability measure on $\Theta$:

$$sp(H) = \sum \{sp(t) : t \in H\} \qquad (24)$$

for all $H \subseteq \Theta$. Property (24) holds for precise hints but does not hold in general (see property 5 above).

So far we have considered the case where there is only one observation from which inferences about $\theta$ is made. Now consider a second observation $x'$ resulting from another realization of the random variable $\omega$. We consider it as the outcome of the random variable $\omega'$ and we assume that $\omega$ and $\omega'$ are independent and identically distributed random variables before any observation is made. Let $\xi'$ denote the observation variable associated with $\omega'$. Of course the set of possible values of $\xi'$ and $\omega'$ are still $X$ and $\Omega$ and the function $f : \Theta \times \Omega \to X$ is still valid when $x'$ is observed. In the same way as the observation $x$ generated the hint $\mathcal{H}(x)$, the observation $x'$ generates the hint

$$\mathcal{H}(x') = (v_{x'}(\Omega), P'', \Gamma_{x'}, \Theta) \qquad (25)$$

where $P''$ is the conditional probability of $P$ given $v_{x'}(\Omega)$. Then the question is: what can be infered about $\theta$ given the hints $\mathcal{H}(x)$ and $\mathcal{H}(x')$ derived from the observations $x$ and $x'$ ?

Under the assumption that the outcome of $(\omega, \omega')$ was $(o, o')$, the observations $x$ and $x'$ imply that $t^* \in \Gamma_x(o)$ and $t^* \in \Gamma_{x'}(o')$ and hence $t^* \in \Gamma_x(o) \cap \Gamma_{x'}(o')$. Therefore, a pair $(o, o')$ such that $\Gamma_x(o) \cap \Gamma_{x'}(o') = \emptyset$ is certainly not the outcome of $(\omega, \omega')$. So, in view of the observations $x$ and $x'$, the set of possible outcomes of $(\omega, \omega')$ is

$$\Omega_{x,x'} = \{(o,o') \in v_x(\Omega) \times v_{x'}(\Omega) : \Gamma_x(o) \cap \Gamma_{x'}(o') \neq \emptyset\}. \qquad (26)$$

So the probability measure $P'P''$ must be conditioned on $\Omega_{x,x'}$, which leads to the new probability space $(\Omega_{x,x'}, P_{x,x'})$. The probability $P_{x,x'}(o, o')$ expresses the chance that the outcome of $(\omega, \omega')$ was $(o, o')$ after observing $x$ and $x'$. Putting things together results in the hint

$$\mathcal{H}(x, x') = (\Omega_{x,x'}, P_{x,x'}, \Gamma_{x,x'}, \Theta) \qquad (27)$$

where

$$\Gamma_{x,x'}(o, o') = \Gamma_x(o) \cap \Gamma_x(o') \qquad (28)$$

for all $(o, o') \in \Omega_{x,x'}$. From this hint degrees of support and plausibility can be computed for any hypothesis $H \subseteq \Theta$ we are interested in. This is done in the same way as with the hint $\mathcal{H}(x)$. For example, the degree of support of $H \subseteq \Theta$ is

$$sp_{x,x'}(H) = P_{x,x'}(u_{x,x'}(H)) \qquad (29)$$

where

$$u_{x,x'}(H) = \{(o,o') \in \Omega_{x,x'} : \Gamma_{x,x'}(o, o') \subseteq H\}. \qquad (30)$$

This way to combine two hints into a new hint is called Dempster's rule of combination. As we have seen, this rule is very logical and is an essential element of the theory of hints and belief functions. Two hints having the same support function are called equivalent. If $\oplus$ denotes the combination operator, i.e.

$$\mathcal{H}(x, x') = \mathcal{H}(x) \oplus \mathcal{H}(x'), \qquad (31)$$

then it can be proved that ⊕ is commutative and associative up to equivalence. This allows us to speak of the combination of more than two hints.

When there is a sequence of observations $x_1, \ldots, x_n$ in a generalized functional model $(f, P)$, the infered knowledge about $\theta$ is represented by the combined hint

$$\mathcal{H}(x_1, \ldots, x_n) = \oplus_{i=1}^{n} \mathcal{H}(x_i). \qquad (32)$$

This hint can then be used to determine degrees of support and plausibility of hypotheses of interest in the usual way. Note that the actual computation of such degrees of support and plausibility can be facilitated by using special computational procedures involving the so-called commonality functions (for more information on this topic, see [4] or [6]).

## 4  Examples

### 4.1  Jessica's Pregnancy Test

Jessica's pregnancy test problem can be analyzed from a functional model perspective. Indeed, from $\omega = \theta \cdot \xi$, it follows that $\theta \cdot \omega = \xi$, which defines the function $f : \Theta \times \Omega \to X$. Actually, the relations $\omega = \theta \cdot \xi$ and $\theta \cdot \omega = \xi$ are mathematically equivalent, both expressing the fact that the test device is trustworthy (i.e. $\omega = +1$) when the observed test result (i.e. the value of $\xi$) is the same as the actual pregnancy status (i.e. the value of $\theta$). If it is assumed that the test result is a correct indicator with probability 0.9 and a wrong indicator with probability 0.1, then it follows that $P(\omega = +1) = 0.9$ and $P(\omega = -1) = 0.1$ because $\omega = +1$ implies $\xi = \theta$ (the test result is a correct indicator) and $\omega = -1$ implies $\xi = -\theta$ (the test result is a wrong indicator). Remark that in the functional model these two probabilities are part of the initial knowledge about the problem, they are not derived from known conditional probabilities as in the fiducial theory. Since Jessica's test result is negative, i.e $\xi = -1$, it follows that $v_{-1}(\Omega) = \Omega$ and

$$\Gamma_{-1}(1) = \{-1\}, \quad \Gamma_{-1}(-1) = \{1\}, \qquad (33)$$

which defines the hint

$$\mathcal{H}(-1) = (\Omega, P, \Gamma_{-1}, \Theta). \qquad (34)$$

This is a precise hint and therefore the associated support function is a regular probability distribution given by

$$sp(-1) = 0.9, \quad sp(1) = 0.1. \qquad (35)$$

So the degree of support that Jessica is pregnant is 0.1 and the degree of support that she is not is 0.9. This is a reasonable result because the reliability of the test device is 0.9 and the test says that Jessica is not pregnant.

In general, if $P(\omega = 1) = p$ and the test is repeated $n$ times, let $k$ denote the number of positive results and $n - k$ the number of negative results. It can be shown that the hint resulting from the combination of the $n$ hints associated with the $n$ observations is again a precise hint whose support function is given by

$$
\begin{aligned}
sp(-1) &= \frac{(1-p)^k p^{n-k}}{p^k(1-p)^{n-k} + (1-p)^k p^{n-k}} \\
sp(+1) &= \frac{p^k(1-p)^{n-k}}{p^k(1-p)^{n-k} + (1-p)^k p^{n-k}}. \qquad (36)
\end{aligned}
$$

### 4.2  Policy Identification (I)

In this example, the following real-world situation is considered. Peter is in room A and has a regular coin in front of him. He flips the coin and observes what shows up: either $H$ or $T$. Then he decides between the following two policies: either tell Paul, who is in room B, what actually showed up on the coin (policy 1) or tell him that the coin showed heads up, regardless of what actually showed up (policy 2). Of course, Paul is unaware of the policy that Peter has chosen. Peter flips the coin $n$ times in total and each time reports to Paul what the chosen policy dictates to tell him. Paul knows that Peter is using the same policy each time the coin is flipped. From the sequence of heads and possibly tails that he receives from Peter, Paul wants to infer information about what policy Peter is using, is it policy 1 or policy 2 ?

To answer this question, Paul builds a generalized functional model of the situation. The parameter space is $\Theta = \{t_1, t_2\}$ where $t_1$ means, say, that Peter is using policy 1 and $t_2$ that he is using policy 2. The observation space is $X = \{H, T\}$ (the set of possible reports to Paul after one flip of the coin) and the sample space of $\omega$ is $\Omega = \{o_1, o_2\}$ where $o_1$ means, say, that heads shows up when the coin is flipped and $o_2$ means that tails shows up when the coin is flipped. Since the coin is a regular one, we have $P(o_1) = P(o_2) = 0.5$. What is the function $f : \Theta \times \Omega \to X$ in this case ? Let's start with $f(t_1, o_1)$. It is clear that under policy 1, if heads shows up then Peter will report $H$ to Paul, which means that $f(t_1, o_1) = H$. It can easily be seen that the function $f$ is given by

$$
\begin{array}{ll}
f(t_1, o_1) = H, & f(t_1, o_2) = T \\
f(t_2, o_1) = H, & f(t_2, o_2) = H.
\end{array} \qquad (37)
$$

First let's analyse the situation when only one value is reported to Paul. If $H$ is reported, then the information that can be infered on $\theta$ is represented by the hint

$$\mathcal{H}(H) = (\Omega, P, \Gamma_H, \Theta) \qquad (38)$$

where

$$\Gamma_H(o_2) = \{t_2\}, \quad \Gamma_H(o_1) = \Theta. \qquad (39)$$

It is important to remark that this hint is not a precise hint because $\Gamma_H(o_1)$ is not a singleton. Indeed, $\Gamma_H(o_1) = \Theta$ because if $H$ is reported and the coin shows heads up then Peter could have used either policy 1 or policy 2. On the other hand, $\Gamma_H(o_2) = \{t_2\}$ because if $H$ is reported and the coin shows tails up then Peter is necessarily using policy 2. The hypothesis that Peter is using policy 2 is supported with strength $P(o_2) = 0.5$ by the observation $H$, whereas there is not support in favor of the fact that Peter is using policy 1. So the support function associated with $\mathcal{H}(H)$ is given by

$$sp_H(t_1) = 0, \quad sp_H(t_2) = 0.5. \quad (40)$$

Since $\mathcal{H}(H)$ is not precise, this support function is not a regular probability measure on $\Theta$: $sp_H(t_1) + sp_H(t_2) < 1$. The plausibility function associated with $\mathcal{H}(H)$ is given by

$$pl_H(t_1) = 0.5, \quad pl_H(t_2) = 1. \quad (41)$$

This means that the degree of compatibility of the hypothesis that Peter is using policy 1 with the observation $H$ is 0.5, whereas nothing is speaking against the hypothesis that he is using policy 2, it is fully compatible with the observation $H$.

If $T$ is reported, then $v_T(\Omega) = \{o_2\}$ because $o_1$ is excluded since it is impossible to observe $T$ if the coin turned heads up. This implies that we have to condition $P$ on $\{o_2\}$, which results in the deterministic probability measure $P'(o_2) = 1$. Since $\Gamma_T(o_2) = \{t_1\}$, which means that if $T$ is reported and the coin turned tails then Peter is using policy 1, it follows that the hint derived from the observation $T$ is

$$\mathcal{H}(T) = (\{o_2\}, P', \Gamma_T, \Theta). \quad (42)$$

Then the corresponding support function is

$$sp_T(t_1) = 1, \quad sp_T(t_2) = 0, \quad (43)$$

which means that Peter is definitely using policy 1. The hint $\mathcal{H}(T)$ is called a deterministic hint on $t_1$ because it is certain that $t_1$ is the correct value of the parameter variable $\theta$.

Now let's analyze the situation where $n$ reports are made. Let $k$ denote the number of heads reports and $n - k$ the number of tails reports (the order in which the heads and tails are reported is irrelevant). The hint on $\theta$ that can be derived from this collection of observations is

$$\mathcal{H}(n, k) = (\oplus_{i=1}^{k} \mathcal{H}(H)) \oplus (\oplus_{i=1}^{n-k} \mathcal{H}(T)). \quad (44)$$

If at least one tails is reported (i.e. $n - k > 0$), then it can easily be proved that $\mathcal{H}(n, k)$ is again a deterministic hint on $t_1$, which means that Peter is using policy 1. This is reasonable because if at least one tails is reported then Peter can't be using policy 2. If no tails is reported, then it can be proved that $\mathcal{H}(n, n)$ is not precise and its support function is given by

$$sp(t_1) = 0, \quad sp(t_2) = 1 - 0.5^n. \quad (45)$$

Note that $sp(t_1) + sp(t_2) < 1$ because $\mathcal{H}(n, n)$ is not precise. The corresponding plausibility function is

$$pl(t_1) = 0.5^n, \quad pl(t_2) = 1. \quad (46)$$

So the hypothesis $t_1$ is becoming less and less plausible as the number of heads reported increases, which is of course very reasonable.

### 4.3 Policy Identification (II)

In this example we consider the following real-world situation. Peter is in room A and has two coins in front of him, one red and one blue. Peter tells Paul, who is in room B, that when the red coin is flipped, the probability that heads shows up is $p_1$ and when the blue coin is flipped, the probability that heads shows up is $p_2$. Peter successively flips the red and blue coins and observes what shows up : either $H$ or $T$ for each coin. Then he decides between the following two policies: either tell Paul what showed up on the red coin (policy 1) or tell him what showed up on the blue coin (policy 2). After the policy is chosen, he informs Paul about what showed up on the coin specified by the policy. Of course, Paul is unaware of the policy that Peter is using. This experiment is repeated $n$ times in total, thereby assuming that the policy Peter is using remains the same throughout all experiments. From the sequence of heads and tails that he receives from Peter, Paul wants to infer information about which policy Peter is using, policy 1 or policy 2 ?

To answer this question, Paul builds a generalized functional model of the situation. The parameter space is $\Theta = \{t_1, t_2\}$ where $t_1$ means, say, that Peter is using policy 1 and $t_2$ that he is using policy 2. The observation space is $X = \{H, T\}$. The sample space is constructed from the two random variables involved in the experiment, namely the result of flipping the red coin (random variable $\omega$) and the result of flipping the blue coin (random variable $\omega'$). The sample space of $\omega$ is $\Omega = \{o_1, o_2\}$ where $o_1$ means that the red coin shows heads up and $o_2$ that it shows tails up. By analogy, we define the sample space $\Omega' = \{o'_1, o'_2\}$ of the random variable $\omega'$. The sample space of $(\omega, \omega')$ is then $\Omega \times \Omega'$ and the corresponding probabilities are

$$\begin{aligned} P(o_1, o'_1) &= p_1 p_2 \\ P(o_1, o'_2) &= p_1(1 - p_2) \\ P(o_2, o'_1) &= (1 - p_1)p_2 \\ P(o_2, o'_2) &= (1 - p_1)(1 - p_2). \end{aligned} \quad (47)$$

What is the function $f : \Theta \times \Omega \times \Omega' \to X$ in this case ? Let's start with $f(t_1, o_1, o'_1)$. It is clear that under policy 1, if the red coin shows heads up then Peter will report $H$ to Paul, which implies that $f(t_1, o_1, o'_1) = H$ and $f(t_1, o_1, o'_2) = H$. It can easily be seen that the complete function $f$ is given by

$$\begin{aligned} f(t_1, o_1, o'_1) &= H, & f(t_1, o_1, o'_2) &= H \\ f(t_1, o_2, o'_1) &= T, & f(t_1, o_2, o'_2) &= T \\ f(t_2, o_1, o'_1) &= H, & f(t_2, o_1, o'_2) &= T \\ f(t_2, o_2, o'_1) &= H, & f(t_2, o_2, o'_2) &= T. \end{aligned} \quad (48)$$

If $H$ is reported, then

$$v_H(\Omega \times \Omega') = \{(o_1, o'_1), (o_1, o'_2), (o_2, o'_1)\} \quad (49)$$

and the conditional distribution of $P$ on $v_H(\Omega \times \Omega')$ is the probability $P'$ given by

$$P'((o_1, o'_1)) = \frac{p_1 p_2}{p_1 + p_2 - p_1 p_2}$$

$$P'((o_1, o_2')) = \frac{p_1(1-p_2)}{p_1 + p_2 - p_1 p_2}$$
$$P'((o_2, o_1')) = \frac{(1-p_1)p_2}{p_1 + p_2 - p_1 p_2}. \quad (50)$$

Also, we have

$$\Gamma_H((o_1, o_1')) = \Theta$$
$$\Gamma_H((o_1, o_2')) = \{t_1\}$$
$$\Gamma_H((o_2, o_1')) = \{t_2\}, \quad (51)$$

which shows that the hint

$$\mathcal{H}(H) = (v_H(\Omega \times \Omega'), P', \Gamma_H, \Theta) \quad (52)$$

is not a precise hint. Its corresponding support function is given by

$$sp_H(t_1) = \frac{p_1(1-p_2)}{p_1 + p_2 - p_1 p_2}$$
$$sp_H(t_2) = \frac{(1-p_1)p_2}{p_1 + p_2 - p_1 p_2}. \quad (53)$$

On the other hand, if $T$ is reported, then

$$v_T(\Omega \times \Omega') = \{(o_1, o_2'), (o_2, o_1'), (o_2, o_2')\} \quad (54)$$

and the conditional distribution of $P$ on $v_T(\Omega \times \Omega')$ is the probability $P'$ given by

$$P'((o_1, o_2')) = \frac{p_1(1-p_2)}{1 - p_1 p_2}$$
$$P'((o_2, o_1')) = \frac{(1-p_1)p_2}{1 - p_1 p_2}$$
$$P'((o_2, o_2')) = \frac{(1-p_1)(1-p_2)}{1 - p_1 p_2}. \quad (55)$$

Also, we have

$$\Gamma_T((o_1, o_2')) = \{t_2\}$$
$$\Gamma_T((o_2, o_1')) = \{t_1\}$$
$$\Gamma_T((o_2, o_2')) = \Theta, \quad (56)$$

which shows that the hint

$$\mathcal{H}(T) = (v_T(\Omega \times \Omega'), P', \Gamma_T, \Theta) \quad (57)$$

is again not a precise hint. Its corresponding support function is given by

$$sp_T(t_1) = \frac{(1-p_1)p_2}{1 - p_1 p_2}$$
$$sp_T(t_2) = \frac{p_1(1-p_2)}{1 - p_1 p_2}. \quad (58)$$

Now let's analyse the situation where $n$ reports are made, $k$ of which beeing $H$ and $n-k$ being $T$ (here again the order in which the heads and tails are reported is irrelevant). Then the hint on $\theta$ induced by the collection of observations is

$$\mathcal{H}(n, k) = (\oplus_{i=1}^{k} \mathcal{H}(H)) \oplus (\oplus_{i=1}^{n-k} \mathcal{H}(T)). \quad (59)$$

It can be proved that this hint is not precise and its associated support function is given by

$$sp(t_1) = \frac{N_1}{D}$$
$$sp(t_2) = \frac{N_2}{D}$$

where

$$N_1 = p_1^k(1-p_1)^{n-k} - (p_1 p_2)^k((1-p_1)(1-p_2))^{n-k}$$
$$N_2 = p_2^k(1-p_2)^{n-k} - (p_1 p_2)^k((1-p_1)(1-p_2))^{n-k}$$

and

$$D = p_1^k(1-p_1)^{n-k} + p_2^k(1-p_2)^{n-k}$$
$$- (p_1 p_2)^k((1-p_1)(1-p_2))^{n-k}.$$

## 5   Prior Information in Functional Models

In a situation where the initial knowledge contains a prior probability distribution $P_0$ on $\theta$, it is easy to include this information into a generalized functional model $(f, P)$. Indeed, this prior information can be viewed as a precise hint on $\theta$ given by

$$\mathcal{H}_0 = (\Theta, P_0, \Gamma, \Theta) \quad (60)$$

where $\Gamma(t) = \{t\}$ for all $\theta \in \Theta$. The support and plausibility functions of this hint is nothing but the prior probability distribution $P_0$. Then this hint $\mathcal{H}_0$ is combined with the hint $\mathcal{H}(x_1, \ldots, x_n)$ by Dempster's rule of combination to obtain a new hint $\mathcal{H}$ expressing the knowledge about $\theta$ derived from all the available information about the problem under investigation:

$$\mathcal{H} = \mathcal{H}_0 \oplus \mathcal{H}(x_1, \ldots, x_n). \quad (61)$$

Then, as usual, degrees of support and plausibility can be determined from $\mathcal{H}$. It turns out that $\mathcal{H}$ is always a precise hint.

On the other hand, from a generalized functional model $(f, P)$ the following conditional distributions of $\xi$ can be defined:

$$Pr(\xi = x | \theta = t) = P(\{o \in \Omega : f(t, o) = x\}) \quad (62)$$

for all $t \in \Theta$. Such conditional probability distributions specify what is called a distribution model. These conditional distributions together with a prior probability $\tilde{P}$ on $\theta$ completely specifies a Bayesian model which is denoted by $(Pr, \tilde{P})$. Then it can be proved that if $P_0 = \tilde{P}$, then the support function associated with $\mathcal{H}$ (which is a probability measure since $\mathcal{H}$ is precise) coincides with posterior distribution of $\theta$ given $x_1, \ldots, x_n$ in the Bayesian model $(Pr, \tilde{P})$. In addition, if $\mathcal{H}(x_1, \ldots, x_n)$ happens to be precise, then its associated support function coincides with the posterior distribution of $\theta$ given $x_1, \ldots, x_n$ in the Bayesian model $(Pr, P^u)$ where $P^u$ is the uniform distribution on $\Theta$.

Finally, let's mention that it is possible to find two different generalized functional models $(f_1, P_1)$ and $(f_2, P_2)$ having the same associated distribution model $Pr(\xi = x|\theta = t)$. This means that generalized functional models have a stronger ability to distinguish between real-world situations than distribution models, which is an important good quality of generalized functional models.

## 6   Conclusion

This paper shows that generalized functional models are very natural in some situations of reasoning under uncertainty. The advantage of this type of models is that they are capable of better representing the information that is really available at the begining of the analysis. For example, there is no need to artificially create prior information on the variable of interest if originally there is none. But if some is available, then it can easily be included into a generalized functional model. The analysis of generalized functional models is inspired from ideas already advocated by R.A. Fisher in his theory of fiducial probability. This analysis naturally leads to the use of support and plausibility functions to express the knowledge about the variable of interest that is generated by the available information. These functions, and not necessarily probability distributions, are the appropriate tools to represent the results of the analysis of a generalized functional model.